%% file: arxiv_submission.tex
\documentclass[wcp]{jmlr}

\usepackage{longtable}
\usepackage[T1]{fontenc}
\usepackage{booktabs}
\usepackage{amsfonts,amssymb,amsmath,bm}
\usepackage[acronym]{glossaries}
\usepackage{mathtools}
\usepackage{multirow}
\usepackage{algorithm}
\usepackage{algpseudocode}
\usepackage{wrapfig}
\usepackage{subcaption}
\usepackage{tcolorbox}

\usepackage{lineno}

\input{commands}

\input{acronyms}

\makeatletter
\let\Ginclude@graphics\@org@Ginclude@graphics 
\makeatother

\jmlrvolume{}
\jmlryear{2024}

\title[Dude:\,Dual Distribution-Aware Context Prompt Learning]{Dude:\,Dual Distribution-Aware Context Prompt Learning For Large Vision-Language Model}

 \author{\Name{Duy M. H. Nguyen$^{*\,1,2,3}$},
  \Name{An T. Le$^{*\,4}$}, \Name{Trung Q. Nguyen$^{2,5}$}, \Name{Nghiem T. Diep$^{2}$},\\ \Name{Tai Nguyen$^{2}$}, \Name{Duy Duong-Tran$^{6,8}$}, \Name{Jan Peters$^{2,4,9}$}, \Name{Li Shen$^{8}$}, \\ \Name{Mathias Niepert$^{1,3}$}, \Name{Daniel Sonntag$^{2,7}$}\\
  \\
  \hspace{-0.07in}\addr$^{1}$University of Stuttgart, \addr$^{2}$German Research Center for Artificial Intelligence,\\
  \hspace{-0.25in}\addr$^{3}$Max Planck Research School for
Intelligent Systems, \addr$^{4}$Technical University of Darmstadt\\\hspace{-0.35in}\addr$^{5}$Technical University of Munich, \addr$^{6}$United States Naval Academy, \addr$^{7}$Oldenburg University\\\hspace{0.12in} \addr$^{8}$University of Pennsylvania, 
\addr$^{9}$Hessian.AI.
\addr$^{*}$Co-equal contribution.}

\begin{document}
\vspace{-4cm}
\maketitle
\vspace{-0.3in}
\begin{abstract}
Prompt learning methods are gaining increasing attention due to their ability to customize large vision-language models to new domains using pre-trained contextual knowledge and minimal training data. However, existing works typically rely on optimizing unified prompt inputs, often struggling with fine-grained classification tasks due to insufficient discriminative attributes. To tackle this, we consider a new framework based on a dual context of 
both \textit{domain-shared} and \textit{class-specific contexts}, where the latter is generated by Large Language Models (LLMs) such as \texttt{GPT}s. Such dual prompt methods enhance the model's feature representation by joining implicit and explicit factors encoded in LLM knowledge.
Moreover, we formulate the Unbalanced Optimal Transport (UOT) theory to quantify the relationships between constructed prompts and visual tokens. Through partial matching, UOT can properly align discrete sets of visual tokens and prompt embeddings under different mass distributions, which is particularly valuable for handling irrelevant or noisy elements, ensuring that the preservation of mass does not restrict transport solutions. Furthermore, UOT's characteristics integrate seamlessly with image augmentation, expanding the training sample pool while maintaining a reasonable distance between perturbed images and prompt inputs. Extensive experiments across few-shot classification and adapter settings substantiate the superiority of our model over current state-of-the-art baselines.
\end{abstract}
\vspace{0.1in}

\section{Introduction}
\begin{figure}[!ht]
    \centering
    \includegraphics[width=\linewidth]{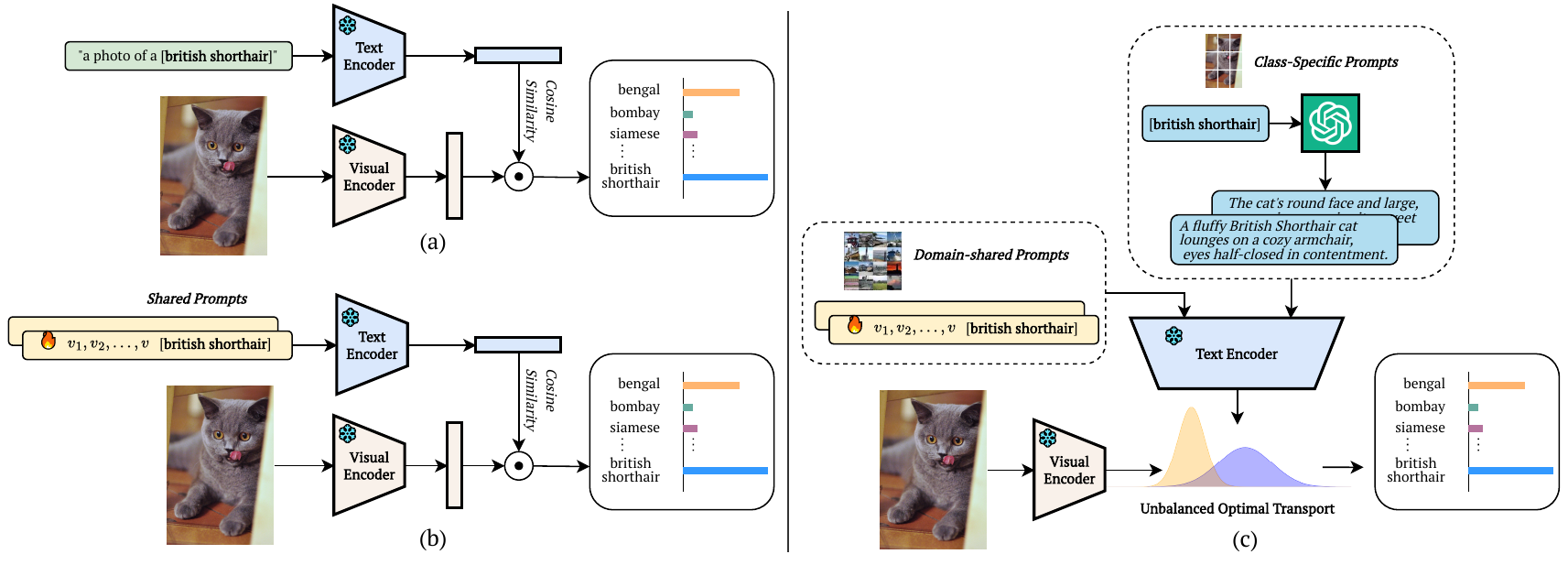}
    \vspace{-0.3in}
    \caption{(a) Zero-shot learning; (b) Shared classes prompt learning; (c) Our method with dual prompts and Unbalanced Optimal Transport (UOT) as the distance between visual tokens and prompt sets.}
    \label{fig:fig1}
    \vspace{-0.2in}
\end{figure}

Recent advancements in vision-language models (VLMs), exemplified by \texttt{CLIP}~\citep{clip}, \texttt{ALIGN}~\citep{align}, or \texttt{Flava}~\citep{flava}, have demonstrated remarkable capabilities in learning comprehensive visual and textual concepts in classification, generation, or recognition. 
During pre-training, these models leverage web-scale image-text pairs to establish aligned representations of images and text through contrastive loss. For instance, through prompts like ``\texttt {A picture of a $\{\textrm{\texttt{label}}\}$}'', VLMs seamlessly transfer their knowledge into downstream applications, employing zero-shot learning by comparing task-specific descriptions with encoded images and texts (Figure \ref{fig:fig1}\,(a)). Such approaches eliminate the need for extensive fine-tuning, underscoring their adaptability and efficiency in various practical scenarios.

However, the effectiveness of these zero-shot capabilities is highly dependent on the quality of the information embedded in the manually created prompts \citep{coop}. While significant improvement can be achieved through prompt engineering \citep{gu2023systematic}, it is time-consuming, requires domain expertise, and has unpredictable performance under domain shifts. As a direct consequence, data-driven approaches (e.g., prompt learning) are introduced to leverage the rich context of additional information for classification. Early efforts considered a single learnable prompt \citep{coop}, image conditional information \citep{zhou2022conditional}, or multiple prompts \citep{lu2022prompt,plot} to implicitly formulate the shared class context, instance-specific context, or context variance, respectively. Another line of work is built on adapter learning, which refines the textual classifier or visual features with a simple learnable feature modulation for each specific task \citep{clip-adapter,tip-adapter,graph-adapter}. 

Despite achieving promising records in few-shot learning, these models face two significant limitations. First, they often employ unified, learnable context prompts shared across all classes, which can overlook the subtle and unique attributes necessary to differentiate closely related or merely indistinguishable categories. As a result, the model may struggle to accurately identify and classify fine details, such as those required in bird classification tasks. Although methods like \texttt{CoOP} \citep{coop} or those leveraging \texttt{GPT} models \citep{naeem2023i2mvformer} adopt class-specific prompts, they are generally limited to zero-shot learning or require substantial labeled data to learn these class-specific prompts to avoid over-fitting due to the increasing of trainable parameters relative to the number of classes.
Second, most prompt-based and adapter models utilize cosine distance between global visual and prompt features to measure affinity, potentially ignoring intricate relationships between image features and textual descriptions (Figure \ref{fig:fig1}\,(b)). This, in turn, falls short of reflecting the fact that different prompts may correspond to distinct image patches. Consequently, the model has difficulty capturing underlying structures and variability within the data that might distinguish closely related objects, resulting in degraded performance when handling fine-grained classification tasks.

In this paper, we propose bridging both \textit{domain-shared} and \textit{class-specific prompts} initialized from \texttt{GPT}, aiming to enrich class-wise descriptions. Learnable domain-shared prompts serve to establish foundational understanding across various categories, ensuring broad applicability and robust generalization capabilities. Concurrently,  
trainable class-specific prompts derived from \texttt{GPT} facilitate specificity by capturing the diverse attributes unique to each object, thereby favoring discriminative abilities in fine-grained distinction tasks. In particular, we learn a shared self-attention mechanism to mitigate the increase in trainable parameters linked with class-specific prompts. This module takes \texttt{GPT} prompts as inputs and generates textual vectors tailored to various categories, which is parameter efficiency while maintaining discriminative power.

Given dual-composed prompts, we compute their textual embedding by feeding into the frozen text encoder (e.g., \texttt{CLIP} text encoder). Then, we express the distance between visual features and prompt embedding as a distance between discrete probability distributions using the unbalanced optimal transport theory \citep{liero2018optimal}. Specifically, we extract all local visual maps for each image rather than a single global representation. This corresponds to a $7 \times 7 $ spatial dimension in the case of ResNet-50 or outputs taken from the multi-head self-attention layer with the Vision Transformer \citep{vit}. The local visual tokens are subsequently aligned to each prompt feature using transport plans computed by solving the UOT and then averaging two distance values to form a final correlation score. Compared with other distances, such as Euclidean or cosine distances,  the UOT can properly align diverse visual features to local prompts and be resilient against
misalignment or feature shift, benefiting from its partial matching flexibility.
This is particularly advantageous when some visual tokens do not have corresponding matches in the prompt sets. Furthermore, these properties make UOT particularly suitable for data augmentation, where input images are augmented with random transformations before alignment with contextual prompts, aiming to enrich training data and enhance the model's generalization capabilities (Figure \ref{fig:fig1}\,(c)). It is worth noting that while a few current works also employ optimal transport between visual and prompt sets, they typically enforce balanced mass preservation constraints between two sets \citep{plot,kim2023zegot}, resulting in sub-optimal mappings in the essence of misalignment or noise outliers.

\vspace{0.1in}
\noindent
In summary, we make the following contributions:
\vspace{-0.05in}
\begin{itemize}
    \item We propose a dual prompt learning approach that captures both unified domain-shared and class-specific contexts, enriched by descriptions generated by \texttt{GPT}.
    \vspace{-0.05in}
    \item The Unbalanced Optimal Transport (UOT) is formulated to capture underlying relationships between local visual tokens and multi-prompt features while being robust to noise and misalignment.
    \vspace{-0.05in}
    \item We assess our performance on fine-grained classification using both few-shot and adapter-based settings and attain state-of-the-art results compared to other leading benchmarks.
\end{itemize}
 
\section{Related Works}
\textbf{Vision-Language Pre-training Algorithms.}
Several approaches are used to \textit{pre-train vision-language models} with large-scale data. They can be divided into reconstruction \citep{hong2021vln,kim2021vilt}, contrastive learning \citep{align,yuan2021multimodal}, graph matching \citep{mh2024lvm,ektefaie2023multimodal}, or fusing several objective losses \citep{kamath2021mdetr,bao2022vlmo}. In this work, we implement data augmentation on input images similarly to contrastive learning but apply it within the context of prompt learning. Here, perturbed images are aligned with prompt embeddings, with features extracted from frozen text encoders. The distance between the augmented visual features and the prompt visual features is then estimated using the Unbalanced Optimal Transport (UOT).

\vspace{0.1in}
\noindent
\textbf{Efficient Transfer Learning.}
Prompt tuning and adapter-based methods are two prominent directions for transferring task-specific knowledge to downstream tasks by tuning minimal parameters. In prompt tuning, early efforts focus on prompt engineering to seek optimal template inputs, aiming at maximum performance of a non-trainable scheme such as a zero-shot \texttt{CLIP} \citep{clip}. Afterward, \texttt{CoOP} \citep{coop} as the pioneer work extends to learnable prompts in few-shot tasks. Following this trend, several works \citep{zhou2022conditional,zhang2022prompting,lu2022prompt,plot} further improve prompt tuning from multiple aspects, such as image-conditional generalization or multiple prompts for diversity.
In contrast, adapter-style approaches customize vision-language models for particular tasks by incorporating lightweight learnable modules on top of the textual and visual feature outputs. For example, \texttt{CLIP-Adapter} \citep{clip-adapter} introduces a trainable bottleneck layer to produce adapted features, which are then merged with the original \texttt{CLIP} outputs via a residual connection. Other advanced adapter-based techniques have also been exploited, such as those employing task-independent strategies \citep{taskres} or leveraging the structural knowledge of data \citep{graph-adapter}.

In contrast to the aforementioned ones, our formulation bridges both domain-shared and specific-class contextual prompts, leveraging \texttt{GPT}-generated descriptions for enhanced model capacities when dealing with fine-grained tasks. We also implement a distance metric between visual tokens and multiple prompts using UOT, which is effective for both \textit{prompt learning} (Section \ref{sec:prompt-learning}) and \textit{adapter learning} (Section \ref{sec:adapter}).

\vspace{0.1in}
\noindent
\textbf{Representation Learning with Optimal Transport.} Optimal Transport (OT) has been widely adopted in machine learning as an objective comparing distributions. Most of the recent successful OT stories define auxiliary training objectives or transformation components~\citep{montesuma2023recent}
with applications in domain adaptation~\citep{courty2014domain, alvarez2020geometric}, Wasserstein GAN~\citep{pmlr-v70-arjovsky17a}, molecular representation learning~\citep{nguyen2024structure}, and robotics planning \citep{le2023accelerating}.
To address real-world scenarios where the mass preservation constraint is too strict, variants such as Unbalanced Optimal Transport (UOT) \citep{liero2018optimal} and entropic regularization \citep{cuturi2013sinkhorn} have been introduced. These extensions have led to the development of entropic UOT \citep{chizat2018scaling}, which combines the flexibility of UOT with the computational advantages of entropic regularization. Till now, UOT has been used in domain adaptation with minibatch training on large datasets \citep{fatras2021unbalanced}, and recently on unsupervised action segmentation \citep{xu2024temporally}, or reactive policy blending in robotics \citep{le2023hierarchical}. In this work, to the best of our knowledge, we first introduce UOT to prompt learning for large-vision language models. 

\section{Methodology} 
\begin{figure}[!ht]
    \centering
    \includegraphics[width=0.75\linewidth]{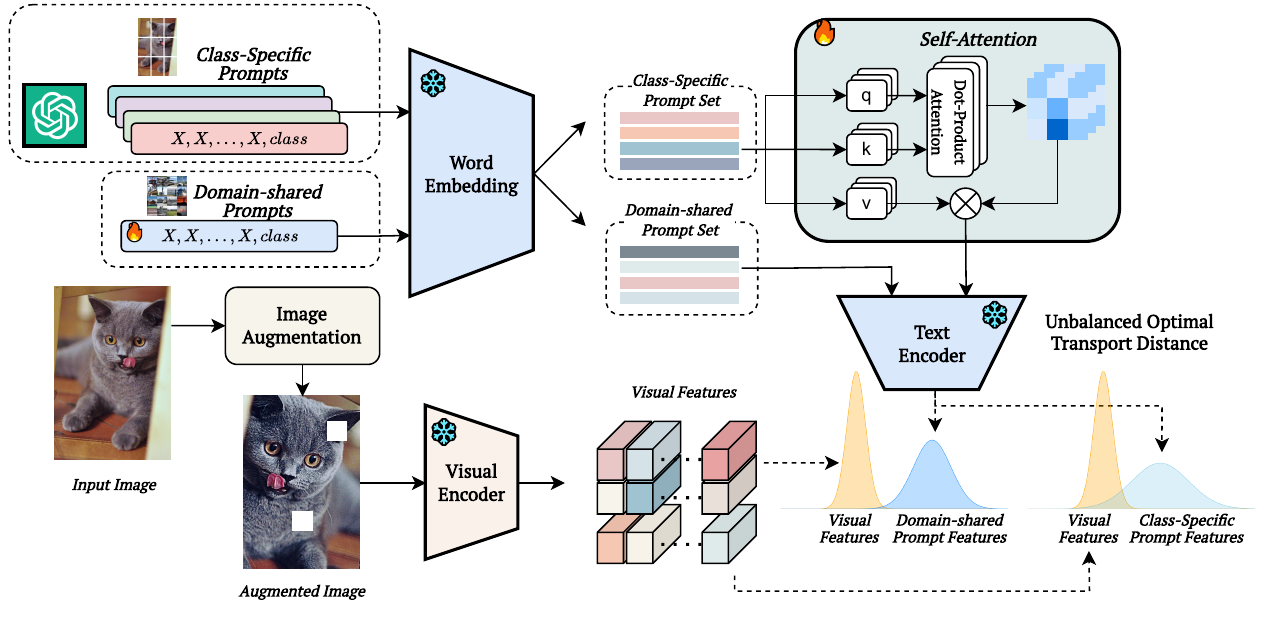}
    \vspace{-0.1in}
    \caption{Overview of the proposed framework. \texttt{CLIP}'s vision and text encoders are frozen, training only domain-shared prompt embeddings and self-attention model.}
    \vspace{-0.1in}
    \label{fig:overview}
\end{figure}
\vspace{-0.1in}
\subsection{Revisit Zero-shot Learning to Single Prompt Learning}
\textbf{Zero-shot Learning.}
Pre-training \texttt{CLIP} \citep{clip} involves learning to match images with their textual descriptions, allowing zero-shot inference on a downstream recognition task by manually designing the prompt template. Let $\vf$ be the feature vector representing an image $\vx \in \gX$, and $\{\vt_i\}_{i=1}^K$ be the prompt tokens generated from an encoder, assuming $\vf, \vt_i$ having the same dimension. $K$ is the total number of classes, and $\vt_i$ is generated from a prompt such as ``\texttt {an image of $\{\textrm{\texttt{label}}\}$}''. The classification likelihood of a class $i$ can be defined as a softmax
\setlength{\abovedisplayskip}{5pt} 
\setlength{\belowdisplayskip}{5pt} 
\begin{equation}
    \sP( c=i \mid \vx) = \frac{\exp( \cos(\vt_i, \vf) / \tau) }{\sum_{j=1}^K \exp( \cos(\vt_j, \vf) /\tau)},
\end{equation}
where $\tau$ is fixed temperature scalar from \texttt{CLIP}, and $\cos(\cdot, \cdot)$ denotes cosine similarity (Figure~\ref{fig:fig1}\,(a)). This differs from traditional classification learning from pre-defined categories in the sense that \texttt{CLIP} leverages natural language descriptions, enabling it to explore a wider range of visual concepts and produce more transferable representations for various tasks.

\vspace{0.1in}
\noindent
\textbf{Prompt Learning.} Zero-shot prediction with fixed prompt features can suffer from domain shift problems. To alleviate,~\cite{zhou2022learning, zhou2022conditional} has demonstrated prompt learning to outperform zero-shot adaptions using manual prompts or linear probe models~\citep{tian2020rethinking}. Specifically, let $\vw$ be the learnable context vector. The learnable prompt is denoted as the concatenation $\vt_i = [\vw_1,\ldots,\vw_{N-1},\vc^i]$, with $\vc^i$ is the word token corresponding to class $i$, having the same dimension as $\vw$. Either same context $\{\vw_k\}_{k=1}^{N-1}$ with all classes or different context $\{\vw_k^i\}_{k=1}^{N-1}$ per class $i$ can be optimized w.r.t. a cross-entropy loss between the labeled target and the prediction
\begin{equation}
    \sP( c=i \mid \vx) = \frac{\exp( \cos(g(\vt_i), \vf) / \tau) }{\sum_{j=1}^K \exp( \cos(g(\vt_j), \vf) /\tau)},
\end{equation}
where $g(\cdot)$ is a text encoder. Recently,~\cite{plot,kim2023zegot} generalize the prompt learning to multi-prompt and multi-visual-features alignment with a distribution-aware OT metric, e.g., the scenarios where many prompts can describe an image and many image regions can be related to a prompt (i.e., many-to-many alignment).
In this work, we look deeper into the matching problem of visual-language alignment, especially the unbalanced problem of visual-language embedding matching described in the next section.

\subsection{Aligning Prompts and Visual Token via Unbalanced Optimal Transport}

Formulating alignment between (multi-)prompt and visual tokens as an OT objective~\citep{plot,kim2023zegot}
with entropic is efficient and scalable. However, we observe that the marginal constraints of OT are restrictive in some settings, e.g., there are many irrelevant image embeddings that are far from true embeddings and hence introduce noises, which should be discouraged entirely (Figure \ref{fig:visualize}\,(Left)). Below, we formally describe the OT and its entropic relaxation, then introduce further relaxation on the marginal constraints, addressing the mentioned problem.
\begin{wrapfigure}{L}{0.52\textwidth}
    \begin{minipage}{0.5\textwidth}
    \vspace{-0.6cm}
\begin{algorithm}[H]
\caption{Solving $\textrm{UOT}_{\lambda}$ in dual form.} \label{alg:uot}
\footnotesize 
\begin{algorithmic}
\State \textbf{Input:} $k = 0$ and $\vu^0 = \vv^0 = \mathbf{0}$.\\ 
\While{not all batch instances converged}{
    \State $\vn^{k} = \mW_{k, k} \mathbf{1}_m$
    \State $\vu^{k+1} = \biggr[ \dfrac{\vu^k}{\lambda} + \log\left(\vn\right) - \log\left(\vn^{k} \right)\biggr] \dfrac{ \lambda \rho_1}{\lambda + \rho_1}$
     \State $\vm^{k} = \mW_{k+1, k}^{\intercal} \mathbf{1}_n$
    \State ${\vv^{k+1} = \biggr[ \dfrac{\vv^k}{\lambda} + \log\left(\vm \right) - \log\left(\vm^{k} \right)\biggr] \dfrac{ \lambda \rho_2}{\lambda + \rho_2}}$
    \State $k \leftarrow k + 1$ \\
}   
\State \textbf{Output:} $\mW^*$.  
\end{algorithmic}
\label{eq:uot_solver}
\end{algorithm}
\vspace{-0.3in}
\end{minipage}
\end{wrapfigure}

\textbf{Notation.} $\mathbf{1}_d$ is the vector of ones in $\sR^d$. The scalar product for vectors and matrices is $x, y \in \sR^d$, $\langle \vx, \vy\rangle = \sum_{i=1}^d \vx_i \vy_i$; and  $\mA, \mB \in \sR^{d \times d}$,  $\langle \mA, \mB\rangle = \sum_{i,j=1}^d \mA_{ij} \mB_{ij}$, respectively. For two histograms $\vn \in \Sigma_n$ and $\vm \in \Sigma_m$ in the simplex $\Sigma_d \defi \{\vx \in \RR^d_+: \vx^\intercal \mathbf{1}_d=1\}$, we define the set
$
  U(\vn, \vm) \defi \{\mW \in \RR_+^{n \times m}\; |\; \mW\mathbf{1}_m=\vn, \mW^\intercal\mathbf{1}_n=\vm\} \, 
$
containing $n \times m$ matrices with row and column sums $\vn$ and $\vm$ respectively. The entropy for $\mA \in U(\vn, \vm)$ is defined as $H(\mA)=-\sum_{i,j=1}^{n, m} a_{ij} \log a_{ij}$. $\widetilde{\mathrm{KL}}(\vw || \vz) = \vw^\intercal\log (\vw \oslash \vz) - \mathbf{1}^\intercal\vw + \mathbf{1}^\intercal\vz$ is the generalized Kullback-Leibler (KL) divergence between two positive vectors $\vw, \vz \in \RR_+^d$ ($\oslash$ is the element-wise division), with the convention $0 \log 0 = 0$. 

Let $\mC \in \sR_+^{n\times m}$ be the positive cost matrix, the OT between $\vn$ and $\vm$ given cost $\mC$ is
$
    \textrm{OT}(\mC) \defi \min_{\mW \in U(\vn, \vm)} \langle  \mW, \mC \rangle.
$
Traditionally, this Kantorovich formulation does not scale well with high dimensions. To address this,~\citet{cuturi2013sinkhorn} proposes to regularize its objective with an entropy term, resulting in the entropic OT
\begin{equation}
\label{eq:otdef}
\textrm{OT}_{\lambda}(\mC) \defi
\textstyle\min_{\mW \in U(\vn, \vm)} \langle \mW, \mC \rangle - \lambda H(\mW),
\end{equation}
which can be solved with the Sinkhorn algorithm~\citep{sinkhorn1967concerning} with complexity of $\Tilde{\gO}(n^2/\epsilon^3)$~\citep{altschuler2017near}, where $\epsilon$ is the approximation error w.r.t. the original $\textrm{OT}(\mC)$. Small $\lambda$ produces fast and biased solutions, or vice versa.

Further relaxing marginal constraints leading to entropic UOT~\citep{chizat2018scaling}
\setlength{\abovedisplayskip}{5pt} 
\setlength{\belowdisplayskip}{5pt} 
\begin{equation}\label{eq:uot}
\textrm{UOT}_{\lambda}(\mC) \defi \min_{\mW \in \sR_+^{n \times m}} \langle \mW, \mC \rangle - \lambda H(\mW) + \rho_1 \widetilde{\mathrm{KL}}(\mW \mathbf{1}_m \;\|\; \vn) + \rho_2\widetilde{\mathrm{KL}}(\mW^\intercal\mathbf{1}_n \;\|\; \vm)
\end{equation}
where now $\vn \in \sR_+^n, \vm \in \sR_+^m$ are arbitrary positive vectors, $\rho_{1,2}$ are the marginal regularization scalars. \Eqref{eq:uot} is well-known as Wasserstein-Fischer-Rao distance on the set of positive Radon measures with entropic regularization \citep{liero2018optimal, sejourne2023unbalanced}, which is desirable as a metric quantifying the alignments of unbalanced embedding distributions on a common latent space.~\citet{pham2020unbalanced} shows that the generalized matrix scaling \Algref{alg:uot}~\citep{chizat2018scaling} solves the dual of \Eqref{eq:uot} 
\begin{equation}\label{eq:dual_uot}
\min_{\vu \in \sR^n,\vv\in \sR^m} \lambda \sum_{i, j = 1}^n \exp \left( \frac{\vu_{i} + \vv_{j} - \mC_{ij}}{\lambda} \right) + \rho_1 \left\langle e^{- \vu/ \rho_1}, \vn \right\rangle + \rho_2 \left\langle e^{- \vv / \rho_2}, \vm \right\rangle,
\end{equation}
with the complexity of $\Tilde{\gO}(n^2/\epsilon)$. Denoting the dual vectors $(\vu^k, \vv^k)$ at iteration $k$, the optimal coupling is computed as $\mW_{i,j} = \textrm{diag}(e^{\vu^i/ \lambda}) \ e^{-\frac{\mC}{\lambda}} \ \textrm{diag}(e^{\vv^j/ \lambda})$. Iterating the Sinkhorn projections (\Algref{alg:uot}) is guaranteed to converge to a fixed point $\mW^*$ (Theorem 4.1 in~\citet{chizat2018scaling}). Note that \Algref{alg:uot} is vectorizable, which is desirable for scaling training with multi-prompt alignments with augmented image patches. We implement the entropic UOT as the alignment distance between a set of image embeddings and a set of word embeddings for each class, with vectorization for minibatch training (i.e., a minibatch of matching sets), described in the next section.

\subsection{Dual Context Prompt Learning}
Despite the scalability and simplicity of using shared multi-prompts \citep{lu2022prompt,plot}, we observe that such a sharing prompt limits its effectiveness in many prompt learning scenarios, such as failing to capture the diverse contexts associated with fine-grained classes. 

\paragraph{Diversifying prompts using LLM.} Due to being pre-trained on extensive corpora, LLMs have acquired substantial common knowledge on a wide range of topics and can serve as external knowledge bases for downstream tasks~\citep{jang2021towards,ke2023continual,razdaibiedina2023progressive}. For each class, we construct a system prompt shown in Figure \ref{fig:prompting} to query the LLM. This aims to obtain image descriptions, providing varied local context information for class \( i \) as a class-specific prompt set \( H_i = \textrm{LLM}(Q_i) \), where \( Q_i \) is the question for class \( i \).


\begin{tcolorbox}[colback=white, colframe=black,
boxrule=0.5pt, 
fontupper=\scriptsize, 
title=Prompting the LLM to generate image descriptions
]

\texttt{System Prompt:}\hspace{-0.2in}
\texttt{
Given the input text indicating the category name of a certain object, your task involves the following steps:
\begin{enumerate}
\setlength\itemsep{0pt}
\item Imagine a scene containing the input object.
\item Generate 4 descriptions about different key appearance features of the input object from the imagined scene, with each description having a maximum of 16 words.
\item Output a JSON object containing the following key: \{``description'': <list of 4 descriptions>\}
\end{enumerate}
}

\end{tcolorbox}
\noindent\begin{minipage}{\textwidth}
\captionsetup{type=figure}

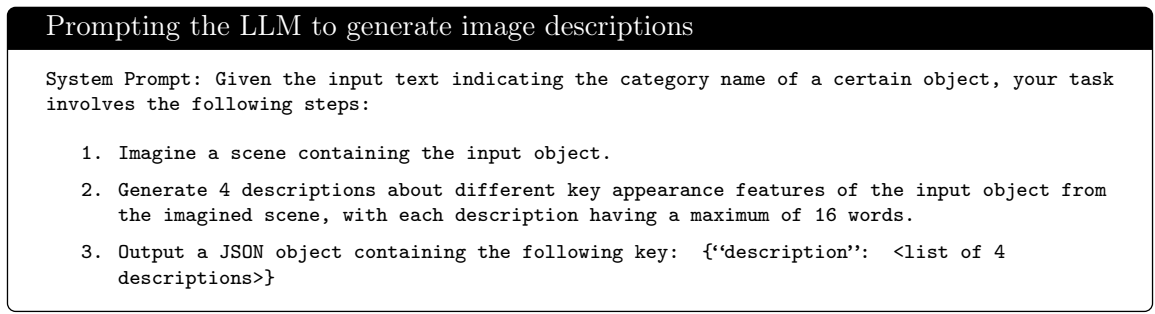
\captionof{figure}{Prompt supplied for the class-specific prompt generation}\label{fig:prompting}
\end{minipage}
For example, in \Figref{fig:fig1}, with the given certain class of ``british shorthair''  in \texttt{OxfordPets} dataset, the response can be \textit{``A fluffy British Shorthair cat lounges on a cozy armchair, eyes half-closed in contentment.''} or\textit{ ``The cat's round face and large, expressive eyes give it a sweet and gentle appearance.''}. Then, the prompts are tokenized using a frozen word embedding into token set \( \{\vw_j^i\}_{j=1}^{N-1} \cup \vc^i \) from \( H_i \).



\paragraph{Self-attention adapter.} Since class-specific prompts can induce exponential increases in token size with an increasing number of classes, we adopt a shared trainable self-attention adapter~\citep{vaswani2017attention} before forwarding the transformed tokens to the frozen text encoder $g(\cdot)$. The self-attention module shown in Figure \ref{fig:overview} is trained on all prompt sets associated with all classes to cope with the exponentially large token number. Intuitively, this module compresses the diverse class contexts with 
$ \textrm{Attention}(\mT) = \text{softmax}\left({\mQ\mK^\intercal}/{\sqrt{d_k}}\right)\mV,
$
where $\mQ = \mT\mW^{Q},\mK=\mT\mW^{K},\mV=\mT\mW^{V} \in \sR^{N \times d_k}$ are the products of the class-specific token matrix $\mT^i = [\vw_1^i, \ldots, \vw_{N-1}^i, \vc^i]^\intercal$ of class $i$ with their associate query, key, value weighting matrices $\mW^{Q}, \mW^{K}, \mW^{V} \in \sR^{d \times d_k}$. 

\paragraph{Image augmentation for diverse visual embeddings.} Data augmentation is a standard technique for combating overfitting~\citep{shorten2019survey}. In the prompt learning setting, we observe that image augmentation generates diverse visual embeddings, preventing overfitting to a subset of local features. Additionally, this technique enhances robustness against common image transformations that happen frequently in practice. Furthermore, the UOT formulation synergizes with data augmentation techniques, as the optimal coupling solution in a balanced problem can constrain the matching to heavily deformed data. For instance, in Figure \ref{fig:overview}, we apply \texttt{random flip}, \texttt{colorjitter}, and \texttt{cutout} transformations on the input images and feed the perturbed outputs to the vision encoder. 
\paragraph{Distribution-aware distance between visual and prompt embedding.}
Let $\mF \in \sR^{M \times d}$ be the image embedding matrix representing $M$ local features from the augmented image $\vx$, $\mG^i_{\textrm{ds}} = g(\mT^i_{\textrm{ds}}) \in \sR^{N \times d}$ be the prompt embedding matrix representing learnable domain-shared prompts, $\mG^i_{\textrm{cs}} = g(\textrm{Attention}(\mT^i_{\textrm{cs}})) \in \sR^{N \times d}$ are class-specific prompt embeddings. We also assume both image embeddings and prompt embeddings lie in the same space $\sR^d$, and are represented by discrete distributions
\setlength{\abovedisplayskip}{3pt} 
\setlength{\belowdisplayskip}{3pt} 
\begin{equation}
    \alpha = \sum_{i=1}^M m_i \delta_{\vf_i},\,\vf_i \in \mF \;\; \beta = \sum_{i=1}^N n_i \delta_{\vg_i},\, \vg_i \in \mG^i,
\end{equation}
where the weights are elements of the marginals $\vm = [m_i]_{i=1}^M, \vn = [n_i]_{i=1}^N$ and can be selected as uniform weights.
The cost between two domains now is defined as $\mC = \mathbf{1}_{n \times m} - \cos(\mF, \mG^i)$, where $\cos(\cdot, \cdot)$ denotes the pairwise cosine similarity between embeddings. Then, the embedding matching objective for class $i$ can be defined as two UOT distances (Eq.~(\ref{eq:uot})), including: (i) class-specific prompts $\textrm{UOT}^i_{\lambda}(\mC_{\textrm{cs}})$ and (ii) domain-shared prompts alignments  $\textrm{UOT}^i_{\lambda}(\mC_{\textrm{ds}})$, respectively.
Given this, the final alignment objective is the weighted sum $d^i = \gamma_{\textrm{cs}}\textrm{UOT}^i_{\lambda}(\mC_{\textrm{cs}}) + \gamma_{\textrm{ds}}\textrm{UOT}^i_{\lambda}(\mC_{\textrm{ds}})$ with the weighting scalars $\gamma_{\textrm{cs}}, \gamma_{\textrm{ds}} > 0$ (Figure~\ref{fig:overview}). The classification likelihood can be written as
\setlength{\abovedisplayskip}{5pt} 
\setlength{\belowdisplayskip}{5pt} 
\begin{equation}\label{eq:likelihood_uot}
\sP( c=i \mid \vx) = \frac{\exp( (1 - d^i) / \tau) }{\sum_{j=1}^K \exp( (1 - d^j) /\tau)}.
\end{equation}
For each inner iteration, we optimize UOT objectives in batches of $K$ classes and fix $\{\mW^*_i\}_{i=1}^K$, then, using Danskin theorem~\citep{danskin1966theory}, we can optimize the prompts the cross-entropy objective~\citep{plot,lu2022prompt,coop}
\begin{equation}\label{eq:lce}
\mathcal{L}_{\textrm{CE}} = - \frac{1}{|\gX|} \sum_{\vx \in \gX} \sum_{i=1}^K y_{i, \vx} \sP( c=i \mid \vx),
\end{equation}
where $\vy_{\vx}$ is the one-hot label for image $\vx$.
\vspace{-0.2in}
\section{Experiment Results}
\subsection{Datasets and Implementation Details}
\paragraph{Datasets.} We conduct \textit{few-shot learning} on five fine-grained datasets, including \texttt{Flowers102} \citep{nilsback2008automated}, \texttt{FGVCAircraft} \citep{maji2013fine}, \texttt{StanfordCars} \citep{krause20133d}, \texttt{OxfordPets}  \citep{parkhi2012cats}, and \texttt{Food101} \citep{bossard2014food}.  
\paragraph{Implementation Details}
Our implementation builds on the CoOp codebase \citep{coop}. We conducted all experiments using \texttt{CLIP} with ViT-B/16 and ResNet-50 backbones. The number of domain-shared and class-specific prompts is chosen to be either $2$ or $4$, depending on the dataset. We use \texttt{ChatGPT} APIs to generate prompts for each class, the system prompt is shown in Figure \ref{fig:prompting}. The final results were averaged over three random seeds (1/2/3) for a fair comparison. We used the Adam optimizer with a learning rate of $2e^{-3}$ and a batch size of 32, running for 50 epochs. We configured self-attention with a single-head output for data efficiency. The UOT problem is solved using the Sinkhorn algorithm, as described in Algorithm \ref{eq:uot_solver}.  We tuned the hyperparameters of $\rho_{1}, \rho_{2}$ in range of $\{\infty, 0.001\rightarrow 0.023\}$ based on validation performance. All experiments were performed on A100 GPUs.

\vspace{-0.1in}
\subsection{Few-shot Learning with Prompt-based Methods}
\label{sec:prompt-learning}
\paragraph{Baselines.} We compare with ten \textit{prompt-based methods} including \texttt{CoOp} \citep{coop}, \texttt{CoCoOp} \citep{zhou2022conditional}, \texttt{DAPT} \citep{dapt}, \texttt{ProGrad} \citep{ProGrad}, \texttt{ProDA} \citep{lu2022prompt}, \texttt{KgCoOp} \citep{kgcoop}, \texttt{RPO} \citep{lee2023read}, \texttt{Plot} \citep{plot}, \texttt{MaPLe} \citep{maple}, and \texttt{PromptSRC} \citep{promptsrc}. Results for baseline are summarized from the literature.  Among these, \texttt{DAPT}, \texttt{PLOT} and \texttt{ProDA} relate to prompt distribution learning, and \texttt{ProDA} or \texttt{PLOT} also adapt multi-prompt mechanisms.

\paragraph{Few-shot learning with $K$-shot labeled images.}
We conduct the few-shot classification on five datasets using $K$-shot labeled images and evaluate trained performance on the testing domain \textit{within the same class space as training ones}. It is worth noting that we freeze both \texttt{CLIP}'s vision and text encoders during training. We only train our prompt embeddings and self-attention model. Table \ref{tab:few-shot} summarizes our results using $4$-shot per class with ViT-B/16. We observe that \textsc{Dude} achieves the best performance in three out of five settings and has a higher average performance than state-of-the-art methods, reaching 76.84$\%$. Notably, on some datasets like \texttt{StanfordCars}, \textsc{Dude} significantly outperforms zero-shot \texttt{CLIP} and single shared-prompt methods like \texttt{CoOp}, with substantial margins of 10.65$\%$ and 3.62$\%$, respectively.
\begin{table}[!ht]
\vspace{-0.15in}
\caption{Few-shot learning compared with \textbf{prompt-based methods}.}
\vspace{-0.1in}
\label{tab:fsl}
\scriptsize
\centering
\resizebox{0.9\textwidth}{!}{
\setlength{\tabcolsep}{2.5pt}
\begin{tabular}{l|ccccccccc|c}
\toprule
              & \texttt{CLIP}  & \texttt{CoOp}  & \texttt{CoCoOp} & \texttt{ProGrad} & \texttt{KgCoOp} & \texttt{MaPLe}  & \texttt{DAPT}  & \texttt{PromptSRC} & \texttt{PLOT}  & \textbf{\textsc{Dude}}\\
\midrule
\texttt{OxfordPets}    & 89.10  & 91.30 & \textbf{93.01}  & 93.21   & 93.20  & 92.05        & 92.17 & 93.23     & 92.55   & 92.01  \\
\texttt{StandfordCars} & 65.70  & 72.73 & 69.10  & 71.75   & 71.98  & 68.70         & 74.40  & 71.83     & 74.93   & \textbf{76.35}  \\
\texttt{Flowers}       & 70.70  & 91.14 & 82.56  & 89.98   & 90.69  & 80.80          & 92.37 & 91.31     & 92.93   & \textbf{94.50}  \\
\texttt{Food101}       & 85.90  & 82.58 & 86.64  & 85.77   & 86.59  & \textbf{86.90}          & 83.60  & 86.06     & 86.46    & 84.90  \\
\texttt{FGVCAircraft}  & 24.90  & 33.18 & 30.87  & 32.93   & 32.47  & 29.03         & 32.47 & 32.80     & 35.29    & \textbf{36.45}  \\
\midrule
Average          & 67.26 & 74.19 & 72.44  & 74.73   & 74.99  & 71.50       & 75.00 & 75.05     & \underline{76.43}   & \textbf{76.84} \\
\bottomrule
\end{tabular}}
\vspace{-1.25em}
\label{tab:few-shot}
\end{table}
\vspace{0.05in}
\paragraph{Base-to-New Class Generalization within Same Domain.}
\begin{table}[t]
\vspace{0.2in}
\caption{Comparison on the \textbf{base-to-new generalization} setting with 16-shot samples.}
\vspace{-0.1in}
\scriptsize
\centering
\label{tab:base2new}
\resizebox{0.9\textwidth}{!}{
\setlength{\tabcolsep}{2.5pt}
\begin{tabular}{lc|ccccccccc|c}
\toprule
  &  & \texttt{CoOp} & \texttt{CoCoOp} & \texttt{DAPT} & \texttt{ProGrad} & \texttt{ProDA} & \texttt{KgCoOp} & \texttt{RPO} & \texttt{PLOT} & \texttt{MaPLe}  & \textbf{\textsc{Dude}}       \\ \hline
\multirow{3}{*}{\texttt{OxfordPets}} & Base                  & 94.47                                                                    & 95.20                                                                      & 95.00                                                                    & 95.07                                                                       & 95.43                                                                     & 94.65                                                                      & 94.63                                                                   & 94.50                                                                                                                                    & \textbf{95.43}                                                                                                                                                                                                          & 94.87 \\
                            & New                   & 96.00                                                                    & 97.69                                                                      & 95.83                                                                    & 97.63                                                                       & \textbf{97.83}                                                                     & 97.76                                                                      & 97.50                                                                   & 96.83                                                                                                                                & 97.76                                                                                                                                                                                                             & 97.16 \\ \hline
\multirow{3}{*}{\texttt{StanfordCars}}       & Base                  & 75.67                                                                    & 70.49                                                                      & 75.80                                                                    & 77.68                                                                       & 74.70                                                                     & 71.76                                                                      & 73.87                                                                   & 79.07                                                                                                                                     & 72.94                                                                                                                                                                                                            & \textbf{80.75} \\
                            & New                   & 67.53                                                                    & 73.59                                                                      & 63.93                                                                    & 68.63                                                                       & 71.20                                                                     & 75.04                                                                      & \textbf{75.53}                                                                   & 74.80                                                                                                                                    & 74.00                                                                                                                                                                                                               & 74.23 \\ \hline
\multirow{3}{*}{\texttt{Flowers}}    & Base                  & 97.27                                                                    & 94.87                                                                      & 96.97                                                                    & 95.54                                                                       & 97.70                                                                     & 95.00                                                                      & 94.13                                                                   & \textbf{97.93}                                                                                                                                    & 95.92                                                                                                                                                                                                              & 97.53 \\
                            & New                   & 67.13                                                                    & 71.75                                                                      & 60.90                                                                    & 71.87                                                                       & 68.68                                                                     & 74.73                                                                      & 76.67                                                                   & 73.53                                                                                                                                   & 72.46                                                                                                                                                                                                         & \textbf{76.73} \\\hline
\multirow{3}{*}{\texttt{Food101}}    & Base                  & 89.37                                                                    & \textbf{90.70}                                                                      & 90.37                                                                    & 90.37                                                                       & 90.30                                                                     & 90.5                                                                       & 90.33                                                                   & 89.80                                                                                                                                     & 90.71                                                                                                                                                                                                                & 90.37 \\
                            & New                   & 88.77                                                                    & 91.29                                                                      & 91.30                                                                    & 89.59                                                                       & 88.57                                                                     & 91.7                                                                       & 90.83                                                                   & 91.37                                                                                                                                 & \textbf{92.05}                                                                                                                                                                                                              & 91.37 \\ \hline
\multirow{3}{*}{\texttt{FGVC-Aircraft}}   & Base                  & 39.67                                                                    & 33.41                                                                      & 39.97                                                                    & 40.54                                                                       & 36.90                                                                     & 36.21                                                                      & 37.33                                                                   & \textbf{42.13}                                                                                                                                     & 37.44                                                                                                                                                                                                             & 42.02 \\
                            & New                   & 31.23                                                                    & 23.71                                                                      & 29.80                                                                    & 27.57                                                                       & 34.13                                                                     & 33.55                                                                      & 34.20                                                                   & 33.73                                                                                                                                    & \textbf{35.61}                                                                                                                                                                                                             & 34.53 \\ \hline
\multirow{3}{*}{Average}    & Base                  & 79.29                                                                    & 76.93                                                                     & 79.62                                                                    & 79.84                                                                       &79.01                                                                     & 77.62                                                                      & 78.06                                                                   & \underline{80.69}                                                                                                                                   & 78.49                                                                                                                                                                                                    & \textbf{81.12} \\
                            & New                   & 70.13                                                                    & 71.61                                                                      & 68.35                                                                    & 71.06                                                                       & 72.12                                                                     & 74.56                                                                       & \underline{74.95}                                                                  & 74.05                                                                                                                                  & 74.38                                                                                                                                                                                                     & \textbf{75.08} \\ \bottomrule                  \end{tabular}}
\vspace{-1.25em}
\label{tab:bsae-new}
\end{table}

We investigate the generalization of prompt tuning by splitting each dataset into two disjoint subsets: \textit{Base} and \textit{New} classes where \textit{Base} categories are utilized for training learnable prompts and \textit{New} categories are used to evaluate performance \citep{lee2023read}. In this setting, we use the ViT-16 \texttt{CLIP} as the base model and train models with 16-shot samples. Table \ref{tab:bsae-new} shows that with increased training examples, all methods improve their performance compared to using only $4$-shot, as seen in Table \ref{tab:few-shot}. Overall, \textsc{Dude} achieves the highest performance in both the \textit{Base} and \textit{New} settings, showcasing its ability to generalize to unseen classes. This capability is attributed to initializing class-specific prompts with external \texttt{GPT} knowledge. Other top-performing baselines, such as \texttt{PLOT} and \texttt{RPO}, also excel by learning multiple prompts and applying regularization to internal feature representations.

\subsection{Few-shot learning with Adapter-based Methods}
\label{sec:adapter}
\begin{figure}[t]
\vspace{0.1in}
    \centering
    \includegraphics[width=0.94\linewidth]{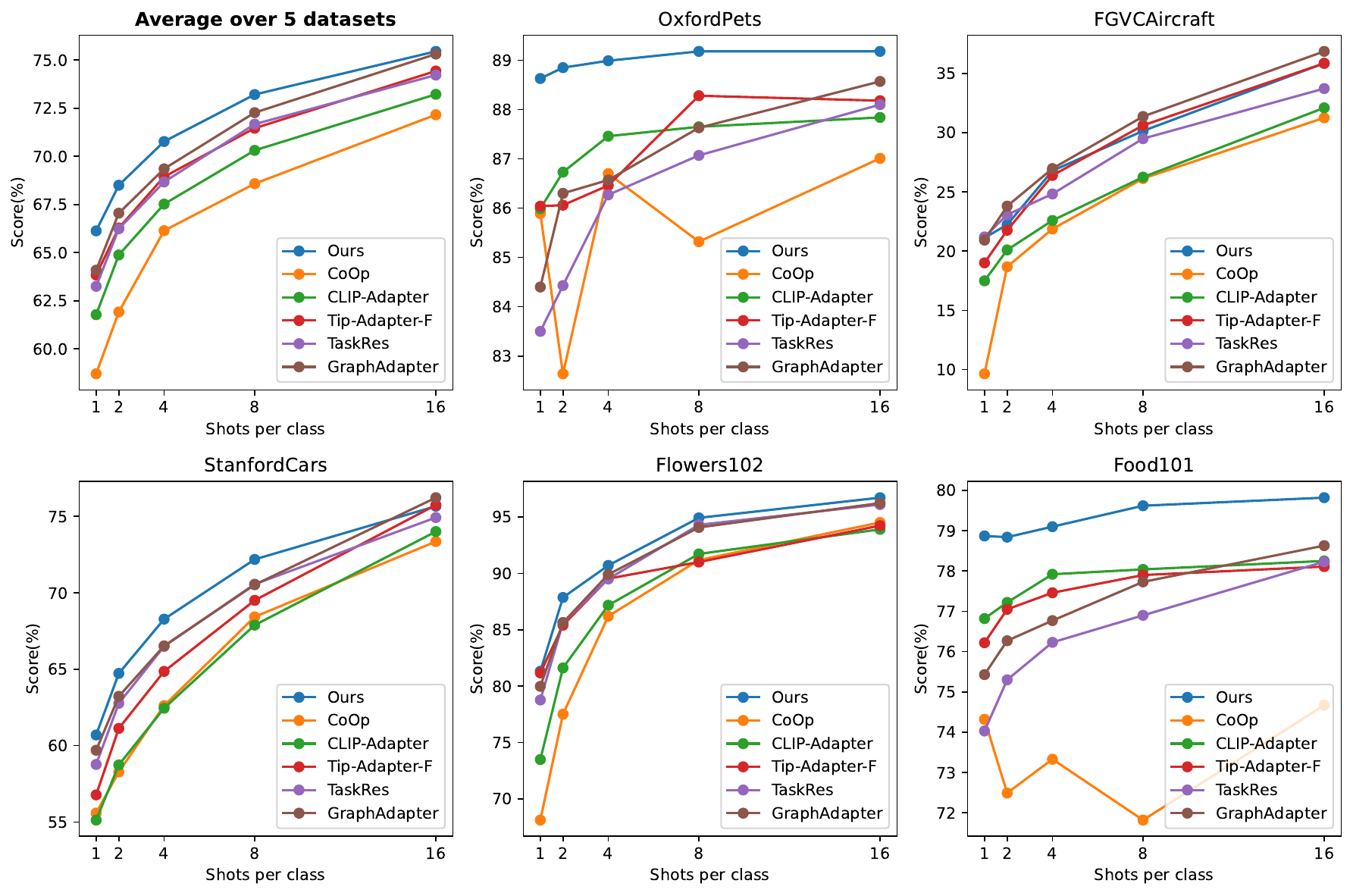}
    \vspace{-0.15in}
    \caption{Few-shot learning results on five datasets with adapter learning. Curves are drawn from $1,2,4, 8, 16$ shots.}
    \vspace{-0.1in}
    \label{fig:adapter-result}
\end{figure}
\paragraph{Settings.} We validate our \textsc{Dude} approach using adapter-based techniques. Rather than optimizing prompt embedding inputs, we focus on training small module networks on the outputs of frozen VML models to adapt to new domains quickly. Our base model uses \texttt{Tip-Adapter} \citep{tip-adapter} with ResNet-50. Specifically, we enhance the original \texttt{Tip-Adapter} by extending from a single learnable linear model to learnable multi-linear models. Additionally, we replace the global embedding in \texttt{Tip-Adapter} with local visual features. Our UOT then reformulates the global embedding-based distance in \texttt{Tip-Adapter} to measure the distance between two distributions.

\paragraph{Baselines.}
We benchmark Adapter-based \textsc{Dude} with the advanced adapter-based baselines involving: \texttt{Clip-Adapter} \citep{clip-adapter}, \texttt{Tip-Adapter} \citep{tip-adapter}, \texttt{TaskRes} \citep{taskres}, and \texttt{GraphAdapter}~\citep{graph-adapter}. All methods are based on the \texttt{CLIP} ResNet-50. 
\paragraph{Results.}
The experimental results are presented in Figure \ref{fig:adapter-result}, proving evidence that our \textsc{Dude} consistently performs better than previous adapter-based methods across 1,\,2,\,4,\,8,\, and 16 shots on four datasets, as well as in average performance. Notably, for datasets such as \texttt{OxfordPets} and \texttt{Food101}, our curves surpass competitive ones by significant margins across all shots. These records, therefore, validate the effectiveness of \textsc{DuDe}, validating its advantage in both prompt learning and adapter cases.

\subsection{Ablation Study}
We implement the following variations to understand the effects of critical components in \textsc{Dude}.
(i) Without learning class-specific context prompts for each class; (ii) without learning domain-shared prompts; (iii) without using \texttt{GPT} to initialize parameters for class-specific prompts, i.e., initialization randomly; (iv) without using unbalanced optimal transport and using standard optimal transport distance; (v) without using shared self-attention to learn class-specific prompt embedding, i.e., each class will initialize separate parameters to train.

\vspace{0.1in}
\noindent
Table \ref{tab:ablation} summarizes the performance of \textsc{Dude} utilizing \texttt{CLIP} ResNet-50 on the \texttt{Food101} and \texttt{OxfordPets} datasets. The results indicate that each component is crucial in achieving optimal performance. Among those, the most important factors include using class-specific context prompts, unbalanced optimal transport as the distance between domains, and the parameter efficiency of shared self-attention for learning per class prompt representations, which avoids amounts of number parameters scaled to the number of categories.

\begin{figure}[!ht]
\centering
\begin{minipage}{.48\textwidth}
  \centering
  \includegraphics[width=.99\linewidth]{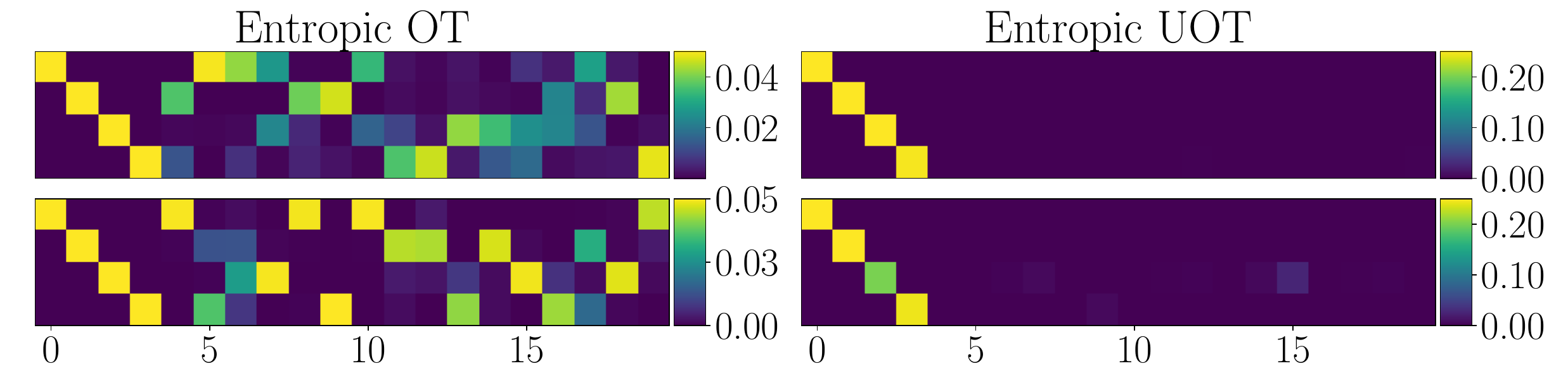}
  \label{fig:OT_vs_UOT}
\end{minipage}%
\begin{minipage}{.255\textwidth}
  \centering
  \includegraphics[width=.99\linewidth]{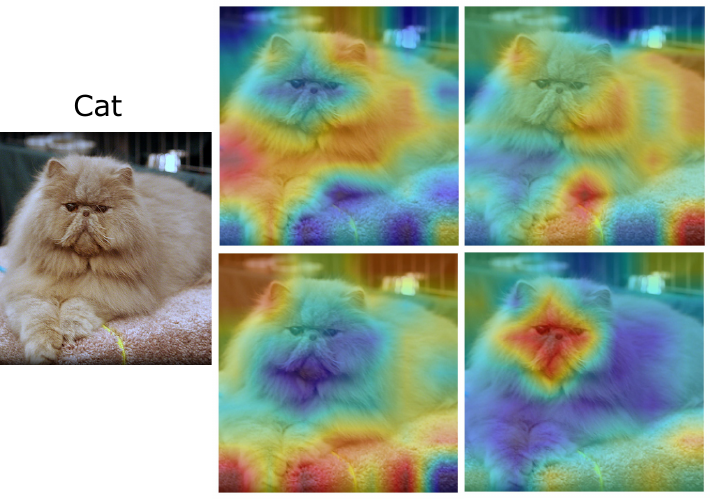}
  \label{fig:cat}
\end{minipage}
\begin{minipage}{.255\textwidth}
  \centering
  \includegraphics[width=.99\linewidth]{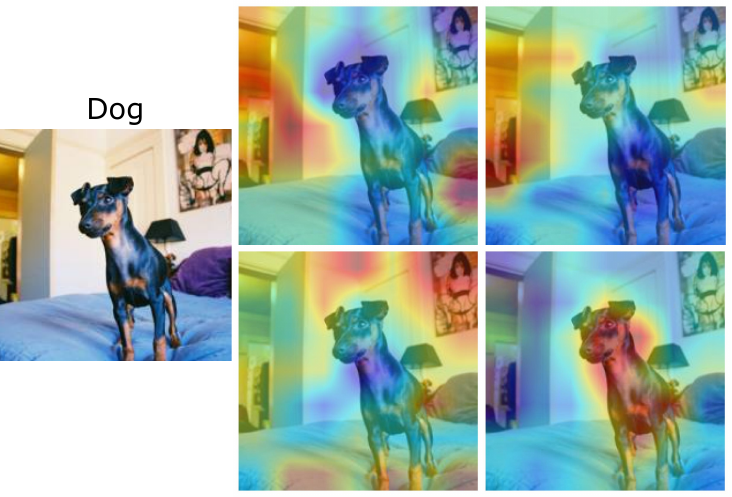}
  \label{fig:dog}
\end{minipage}
\vspace{-0.2in}
\caption{(\textbf{Left}) Comparison between Balanced OT and Unbalanced OT on the \texttt{Food101} (top) and the \texttt{OxfordPets} dataset (bottom); (\textbf{Right}) heatmaps of optimal transport plan related to each of class-specific context prompts learned from \texttt{GPT} on two examples of \texttt{Cat} and \texttt{Dog}.\vspace{-0.15in}}
\label{fig:visualize}
\end{figure}
\vspace{-0.2in}
\subsection{Visualization}
\paragraph{Transport Mapping from Balanced and Unbalanced OT.}
In~\Figref{fig:visualize} (left),  we provide an intuitive example of the output differences in optimal coupling of entropic OT and entropic UOT under outliers. In particular, we show multi-prompt alignment between $4$ prompts and $20$ images where only $4$ images matched with prompts; others are negative samples. In the UOT setting, we set $\rho_1 \rightarrow \infty, \rho_2 = 0.04, \lambda = 0.01$ for conserving source marginal while relaxing target marginal. Clearly, the optimal coupling of entropic OT is blurry, thus introducing matching noises, while entropic UOT destroys noisy couplings and produces sharper matching. Intuitively, the total mass is conserved between the source and target distributions in entropic OT. However, this marginal constraint is restrictive in multi-prompt alignment problems where several word embeddings might not properly correspond to local visual ones, especially under data augmentation.

\vspace{-0.05in}
\begin{figure}[!ht]
\begin{minipage}[t]{0.6\linewidth}
    \centering
    \vspace{-0.05in}
    \captionof{table}{{\textbf{Ablation studies on few-shot recognition}: \texttt{CSC Prompt}: Class-specific context prompts for each class. \texttt{SC Prompt}: Domain-shared class prompts. \texttt{Self Att}: Shared Attention for all prompts}}
    \label{tab:ablation}
    \vspace{-0.1in}
    \resizebox{\textwidth}{!}{
    \setlength{\tabcolsep}{2.pt}
    \begin{tabular}{ccclllclllclllclllclll}
    \toprule
    \textbf{Dataset} &
      \textbf{Setting} &
      \textbf{1 shot} &
      \textbf{2 shot} &
      \textbf{4 shot} &
      \textbf{8 shot} &
      \textbf{16 shot} \\ \midrule
    \multirow{6}{*}{\texttt{Food101}} &
      \textbf{Our (full)} &
      \textbf{77.8} &
      \textbf{77.8} &
      \textbf{77.9} &
      \textbf{78.5} &
      \textbf{78.7} \\
     & w/o CSC Prompt       & 77.6 & 77.8  & 77.1  & 75.4 & 77.1  \\
     & w/o SC Prompt        & 75.4 & 77.1 & 77.3 & 77.8  & 78.4 \\
     & w/o GPT init         & 76.5 & 77.4 & 77.7 & 77.3 & 78.1 \\
     & w/o UOT    & 75.7  & 76.8 & 77.2 & 77.6 & 78.3 \\
     & w/o Self Att & 61.8 & 68.0 & 70.8  & 74.1 & 75.7  \\ \midrule
     \multirow{6}{*}{\texttt{OxfordPets}} &
      \textbf{Our (full)} &
      \textbf{87.5} &
      \textbf{87.5} &
      \textbf{88.1} &
      \textbf{88.9} &
      \textbf{88.4} \\
     & w/o CSC Prompt       & 87.3  & 86.9  & 88.5 & 87.4  & 87.1 \\
     & w/o SC Prompt        & 84.3 & 87.0 & 87.5 & 87.7  & 88.1 \\
     & w/o GPT init         & 86.5 & 87.2 & 87.5 & 88.2  & 87.8  \\
     & w/o  UOT    & 85.7 & 86.7 & 86.9  & 87.4  & 87.9 \\
     & w/o Self Att & 82.5 & 83.1 & 85.5  & 85.8 & 87.6 \\ \bottomrule
    \end{tabular}
    }
\end{minipage}%
\hfill
\begin{minipage}[t]{0.38\textwidth}
\paragraph{Learnable Class-Specific Context Prompt.} Figure \ref{fig:visualize} (right) presents the heatmap of four learnable prompts for each class. The UOT distance between each prompt embedding and visual local features is computed, illustrating correlations from transport plans. It is intuition to observe that each prompt targets distinct sub-regions of the image, covering object characteristics and relevant background. Such properties, therefore, may offer better guidance than a single shared class prompt, resulting in improved predictions.
\end{minipage}
\end{figure}


\vspace{-0.4in}
\section{Conclusion}
This paper demonstrated that a large vision-language model like \texttt{CLIP} can be transformed into a data-efficient learner through prompt learning, utilizing a unified context and class-specific context initialized from the \texttt{GPT} model. Additionally, framing the distance between visual tokens and prompt features as an unbalanced optimal transport problem is essential for capturing misalignments and outliers between the two domains. This approach, combined with data augmentation to increase training samples, significantly enhances the model's few-shot learning abilities. Our results with prompt and adapter-based settings indicate substantial improvements over several competitive approaches. For future work, we propose to (i) test our framework on various types of adapter-based learning to validate its generalization capabilities and (ii) extend the method to vision-language model families trained with the autoregressive setting, such as LLAVA \citep{liu2024visual}. This is particularly challenging since the learned embedding space structures in autoregressive models differ from those in \texttt{CLIP}, which is trained using a contrastive function.
\bibliography{acml24.bbl,ot.bbl}






\end{document}

%% file: commands.tex
\renewcommand{\H}[2][]{
\ifthenelse {\equal{#1}{}}
{\mathbb{H}\left[#2\right]}
{\mathbb{H}_{#1}\left[#2\right]}}

\newcommand{\E}[2][]{
\ifthenelse {\equal{#1}{}}
{\mathbb{E}\left[#2\right]}
{\mathbb{E}_{#1}\left[#2\right]}}

\def\Figref#1{Figure~\ref{#1}}

\def\Eqref#1{Equation~\ref{#1}}

\def\Algref#1{Algorithm~\ref{#1}}

\def\1{\bm{1}}

\def\RR{\mathbb{R}}

\def\vc{{\bm{c}}}

\def\vf{{\bm{f}}}
\def\vg{{\bm{g}}}

\def\vm{{\bm{m}}}
\def\vn{{\bm{n}}}

\def\vt{{\bm{t}}}
\def\vu{{\bm{u}}}
\def\vv{{\bm{v}}}
\def\vw{{\bm{w}}}
\def\vx{{\bm{x}}}
\def\vy{{\bm{y}}}

\def\vz{{\bm{z}}}

\def\mA{{\bm{A}}}
\def\mB{{\bm{B}}}
\def\mC{{\bm{C}}}

\def\mF{{\bm{F}}}
\def\mG{{\bm{G}}}

\def\mK{{\bm{K}}}

\def\mQ{{\bm{Q}}}

\def\mT{{\bm{T}}}

\def\mV{{\bm{V}}}
\def\mW{{\bm{W}}}

\DeclareMathAlphabet{\mathsfit}{\encodingdefault}{\sfdefault}{m}{sl}
\SetMathAlphabet{\mathsfit}{bold}{\encodingdefault}{\sfdefault}{bx}{n}

\def\gO{{\mathcal{O}}}

\def\gX{{\mathcal{X}}}

\def\sP{{\mathbb{P}}}

\def\sR{{\mathbb{R}}}

\newcommand{\defi}{\;{:=}\;}

\makeatletter
\newcommand{\StatexIndent}[1][3]{%
  \setlength\@tempdima{\algorithmicindent}%
  \Statex\hskip\dimexpr#1\@tempdima\relax}
\makeatother

%% file: acronyms.tex
\newacronym{kld}{KL divergence}{Kullback–Leibler Divergence}
\newacronym{kl}{KL}{Kullback–Leibler}
\newacronym{map}{MAP}{maximum a Posteriori}

\newacronym{gp}{GP}{Gaussian Process}
\newacronym{ot}{OT}{Optimal Transport}
\newacronym{uot}{UOT}{Unbalanced Optimal Transport}
\newacronym{eot}{EOT}{Entropic Optimal Transport}